\documentclass{article}

\PassOptionsToPackage{svgnames, dvipsnames, table}{xcolor}
\PassOptionsToPackage{numbers, compress}{natbib}

\usepackage[preprint]{neurips_2025}

\usepackage{microtype}
\usepackage{hyperref}
\usepackage{url}
\usepackage{booktabs}

\usepackage{lineno}

\usepackage{xspace}
\usepackage{graphicx}
\usepackage{marvosym}

\usepackage{amsmath}
\usepackage{subcaption}
\usepackage{graphicx}
\usepackage{changes}
\usepackage{amsmath,mathtools}
\usepackage{changes}
\usepackage{pifont}

\usepackage{lipsum}
\usepackage{colortbl}
\usepackage{wrapfig}
\usepackage{xspace}
\usepackage{multirow}
\usepackage{enumitem}

\usepackage{pifont}
\usepackage{listings}
\usepackage{fontawesome}

\usepackage{wrapfig}
\usepackage{caption}
\usepackage{mdframed}

\usepackage{makecell}
\usepackage[utf8]{inputenc} 
\usepackage[T1]{fontenc}    
\usepackage{hyperref}       
\usepackage{url}            
\usepackage{booktabs}       
\usepackage{amsfonts}       
\usepackage{nicefrac}       
\usepackage{microtype}      
\usepackage{listings}
\usepackage{tcolorbox}
\tcbuselibrary{listings}
\usepackage{graphicx}
\usepackage{amsmath}
\usepackage{verbatim}
\tcbuselibrary{skins} 
\usepackage{amsfonts} 
\usepackage{mathrsfs} 
\usepackage{algorithm}
\usepackage[noend]{algpseudocode} 
\usepackage{geometry}
\usepackage{tikz}
\usetikzlibrary{shapes.geometric, arrows.meta, positioning, fit, calc}
\usepackage{wrapfig}
\usepackage{multirow}
\usepackage{pgfplots}
\pgfplotsset{compat=1.17} 
\usepackage{tcolorbox}
\usepackage{subcaption}
\tcbuselibrary{breakable} 
\tcbuselibrary{skins}

\definecolor{mypurple}{HTML}{800080} 

\newcommand{\ourmethod}{\emph{FlexiVe} }
\newcommand{\ourpipeline}{\emph{Solve-Detect-Verify} }
\newsavebox{\rightimagebox}

\algnewcommand\And{\textbf{and}}
\algnewcommand\Or{\textbf{or}}
\algnewcommand\Not{\textbf{not}}

\expandafter\def\expandafter\normalsize\expandafter{%
    \normalsize%
    \setlength\abovedisplayskip{0pt}%
    \setlength\belowdisplayskip{0pt}%
    \setlength\abovedisplayshortskip{-8pt}%
    \setlength\belowdisplayshortskip{0pt}%
}
\usepackage{titlesec}

\titlespacing*{\subsection}{0pt}{1pt}{1pt}
\titlespacing*{\section}{0pt}{2pt}{2pt}
\titlespacing*{\paragraph}{0pt}{0pt}{2pt}

\newcommand{\Msolve}{M_{\text{solve}}}
\newcommand{\Khesitation}{\mathcal{K}_{\text{hesitation}}}
\newcommand{\PromptComplete}[1]{\text{Prompt}_{\text{complete}}(#1)}

\newcommand{\EOS}{\text{EOS}}
\definecolor{myPaletteYellowOrange}{HTML}{FF9F00}
\definecolor{myPaletteOrangeRed}{HTML}{F4631E}
\definecolor{myPaletteRed}{HTML}{CB0404}
\definecolor{myPaletteTeal}{HTML}{309898}

\usepackage{tcolorbox}
\tcbuselibrary{breakable} 
\tcbuselibrary{skins}    
\usepackage{lipsum}

\newtcolorbox{customtakeaway}[1][]{
  enhanced,
  breakable, 
  attach boxed title to top left={yshift=-2.5mm, xshift=3mm},
  colback=white,
  colframe=black,
  fonttitle=\bfseries\sffamily,
  colbacktitle=black,
  coltitle=white,
  boxrule=0.8pt,
  arc=2mm,
  boxed title style={
    colframe=black,
    boxrule=0.8pt,
    arc=1.5mm,
    bottomrule=0pt,
    sharp corners=south,
    left=3mm,
    right=3mm,
    top=0.5mm,
    bottom=0.5mm
  },
  top=2mm, 
  left=4mm,
  right=4mm,
  bottom=4mm,
  before skip=2ex,
  after skip=2ex,
  fuzzy shadow={0mm}{0.5mm}{0.25mm}{0.1mm}{black!30!white}, 
  #1
}

\title{\ourpipeline: Inference-Time Scaling with Flexible Generative Verifier}

\author{
  \textbf{Jianyuan Zhong\thanks{\ \ Equal contribution.}}, 
  \textbf{Zeju Li\footnotemark[1]}, 
  \textbf{Zhijian Xu}, 
  \textbf{Xiangyu Wen}, 
  \textbf{Kezhi Li},
  \textbf{Qiang Xu\thanks{\ \ Corresponding author.}} 
  \\
  The Chinese University of Hong Kong \\
  \texttt{\{jyzhong, zjli24, zjxu21, xywen22, kzli24, qxu\}@cse.cuhk.edu.hk}
}

\begin{document}

\maketitle

\vspace{-20pt}
\begin{abstract}
Large Language Model (LLM) reasoning for complex tasks inherently involves a trade-off between solution accuracy and computational efficiency. The subsequent step of verification, while intended to improve performance, further complicates this landscape by introducing its own challenging trade-off: sophisticated Generative Reward Models (GenRMs) can be computationally prohibitive if naively integrated with LLMs at test-time, while simpler, faster methods may lack reliability.  To overcome these challenges, we introduce \ourmethod, a novel generative verifier that flexibly balances computational resources between rapid, reliable ``fast thinking'' and meticulous ``slow thinking'' using a Flexible Allocation of Verification Budget strategy. We further propose the \ourpipeline pipeline, an efficient inference-time scaling framework that intelligently integrates \ourmethod, proactively identifying solution completion points to trigger targeted verification and provide focused solver feedback. Experiments show \ourmethod achieves superior accuracy in pinpointing errors within reasoning traces on ProcessBench. Furthermore, on challenging mathematical reasoning benchmarks (AIME 2024, AIME 2025, and CNMO), our full approach outperforms baselines like self-consistency in reasoning accuracy and inference efficiency. Our system offers a scalable and effective solution to enhance LLM reasoning at test time.
\end{abstract}


\begin{figure*}[h!]
    \vspace{-15pt}
    \centering
    \includegraphics[width=0.95\textwidth]{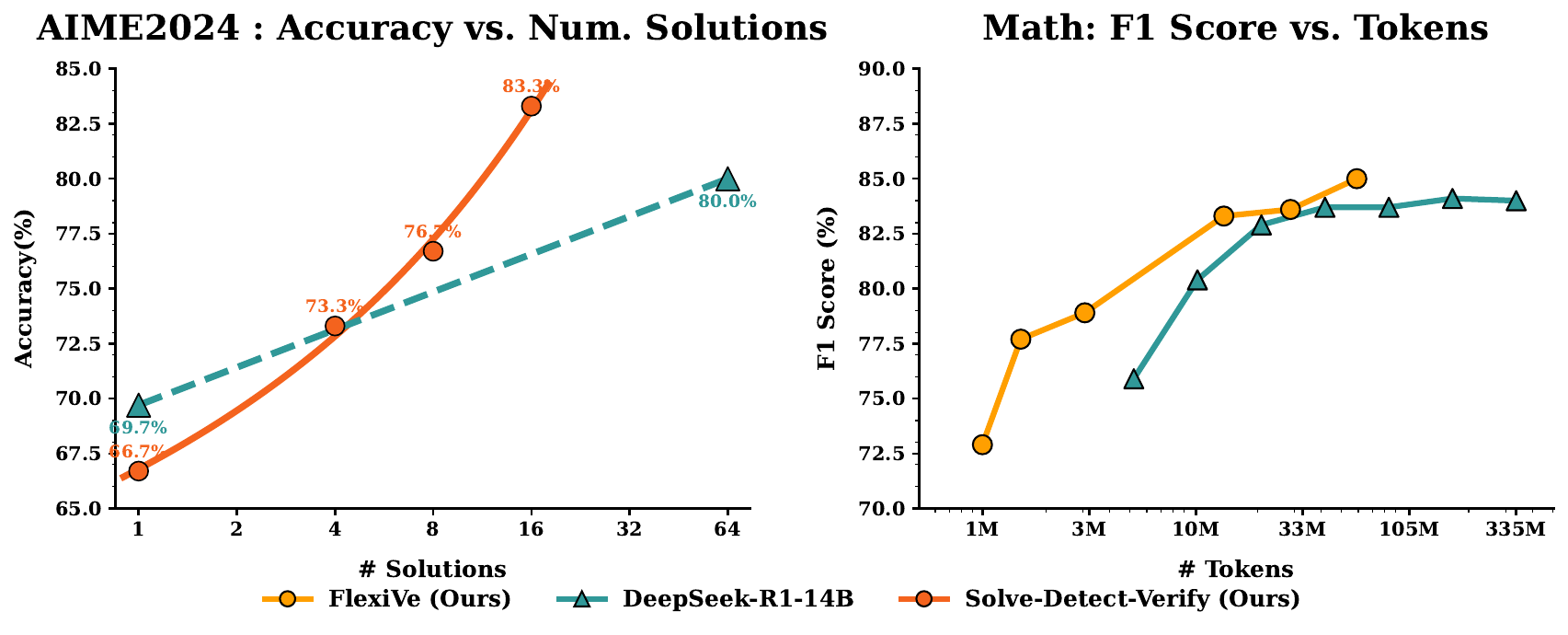}
    \caption{Performance Scaling Analysis. 
    (\textbf{Left}) On the AIME2024 benchmark, our inference-time scaling framework, \ourpipeline, achieves higher accuracy while requiring approximately \textbf{4x fewer solutions} compared to baseline approaches. Since \texttt{DeepSeek-R1-Distill-Qwen-14B} does not report performance from $k=2...32$, we connect two dots with a dotted straight line.
    (\textbf{Right}) On the Math benchmark, our verifier \ourmethod (specifically with the Flex@8 configuration) attains a higher F1 score while generating approximately \textbf{3x fewer tokens} than the baseline.}
    \label{fig:combined_scaling_highlighted}
\end{figure*}


\section{Introduction}
Recent advances in Large Language Models (LLMs) have significantly enhanced their capabilities in tackling complex reasoning tasks, primarily through the explicit generation of step-by-step reasoning traces~\cite{wei2022chainofthought,kojima2022large}. This shift towards deeper, more analytical "System 2" processes~\cite{kahneman2011thinking,li2025system,openai2024reasoningmodels,shao2024deepseekmath,google2025gemini2.5pro}, while crucial for improving solution accuracy, inherently presents a fundamental trade-off with computational efficiency. Models often produce verbose reasoning, including redundant steps or ``overthinking''~\cite{chen2024overthinking}, where extensive intermediate computations needed for higher accuracy incur substantial costs, sometimes for only marginal gains. This landscape highlights the ongoing challenge of balancing accuracy and efficiency in LLM reasoning, necessitating more sophisticated mechanisms for both generating solutions and verifying their correctness.

\begin{figure}[h!]
    \centering
    \includegraphics[width=0.75\textwidth]{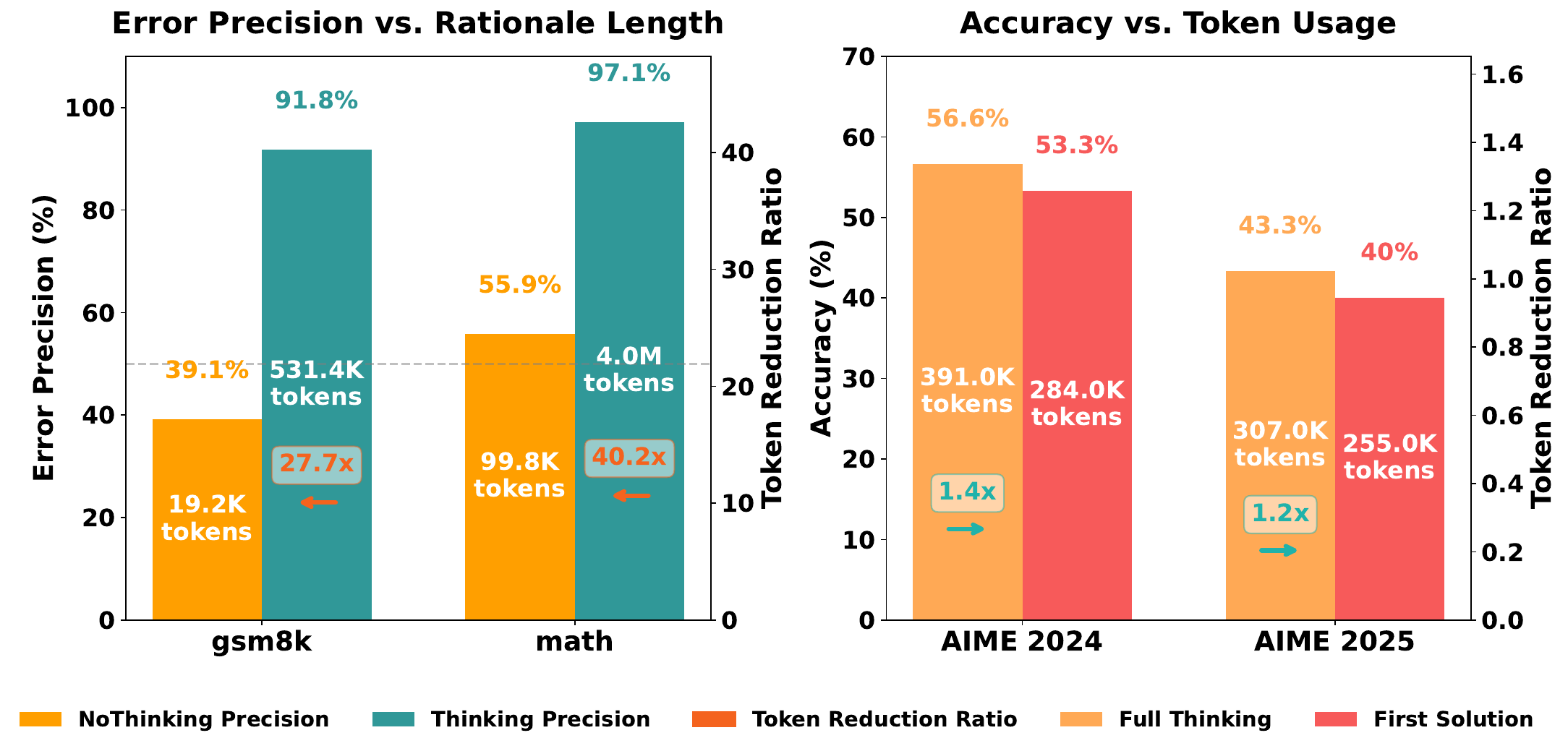}
    \caption{Empirical motivation for efficient verification and generation strategies. \textbf{(Left)} Comparison of error precision and token usage between \textit{NoThinking} and \textit{Thinking} verification on GSM8K and Math (ProcessBench). While \textit{NoThinking} significantly reduces tokens, its error precision is substantially lower, sugguesting high false positive rate. \textbf{(Right)} Accuracy and token usage comparison between generating a full solution (\textit{Full Thinking}) and halting generation early upon detecting a complete intermediate solution (\textit{First Solution}) on AIME 2024 and AIME 2025. Early detection offers significant token reduction with comparable accuracy.}
    \label{fig:motivation}
    \vspace{-15pt}
\end{figure}

The need to ensure the reliability of these reasoning traces through verification further complicates the aforementioned accuracy-efficiency balance~\cite{chen2025sets}. While robust verification is crucial for enhancing LLM capabilities, existing methods introduce their own challenging trade-offs. For example, Generative Reward Models (GenRMs) promise detailed step-level feedback~\cite{liu2025inferencetimescalinggeneralistreward,zhang2025generativeverifiersrewardmodeling}, but often at the cost of significant computational overhead or naive and expensive integration~\cite{singhi2025solveverifycomputeoptimalproblem}. Conversely, highly token-efficient mechanisms like ``NoThinking''~\cite{ma2025nothinking}, when adapted for \textit{verification}, can achieve substantial token reduction (e.g., 27x-40x fewer tokens, see Figure~\ref{fig:motivation}) but suffer a severe drop in error precision (e.g., to 39-56\% on mathematical benchmarks), leading to unreliable judgments. This underscores the critical demand for verifiers that can effectively reconcile speed with high reliability.

This initial efficiency challenge within the reasoning process itself is further exacerbated when LLMs exhibit prolonged ``self-correction'' behavior. Models frequently generate hesitation words or phrases (e.g., ``hmm'', ``let me double check'') and redundant internal verification steps even after a correct intermediate solution might have been implicitly reached~\cite{chen2024overthinking}. This continued generation, as models ``overthink'' a problem, incurs substantial computational costs for little to no gain in final accuracy. An effective system must therefore also address these redundancies by intelligently discerning when a solution is likely complete.

This complex interplay of trade-offs in reasoning and verification reveals a clear methodological gap: there is a pressing need for (1) a flexible verifier that can dynamically adapt its computational effort to the complexity of the verification task, balancing inference speed with accuracy, and (2) an intelligent inference-time pipeline that strategically deploys such a verifier and streamlines the overall reasoning process by curtailing unnecessary computation.
To address these compounded challenges, we introduce two main contributions:

We propose \ourmethod{} (\textbf{Flexi}ble Generative \textbf{Ve}rifier), a novel generative verification method that dynamically adjusts its computational resources. \ourmethod{} employs a rapid, resource-efficient ``fast thinking'' mode, optimized for concise error diagnosis through techniques like Group Relative Policy Optimization (GRPO)~\cite{shao2024deepseekmath,yu2024grpo}, and a thorough, computationally-intensive ``slow thinking'' mode. The transition between these modes is governed by a Flexible Allocation of Verification Budget strategy; this strategy first uses efficient, parallelizable assessments of the entire reasoning trace to gauge verification difficulty, escalating to deeper analysis only when initial consensus is low, thereby allowing \ourmethod{} to analyze entire reasoning traces efficiently and pinpoint errors with high precision, unlike verifiers that operate on a per-step basis~\cite{zhong2025dyvethinkingfastslow}.

To effectively leverage \ourmethod{}, we introduce the \ourpipeline{} pipeline, a novel inference-time scaling framework. This pipeline features a lightweight assessment mechanism that continuously monitors the solver LLM's reasoning trace for cues of solution completeness. Upon detecting a potentially complete solution, the pipeline pauses generation and invokes \ourmethod{} for targeted verification. If validated, the solution is finalized, saving further computation. If errors are found, \ourmethod{}'s focused feedback guides the solver towards refining its reasoning.

Extensive experiments validate both contributions. \ourmethod{} demonstrates superior accuracy in identifying and pinpointing errors within reasoning traces compared to existing verification methods on benchmarks like ProcessBench~\cite{zheng2024processbenchidentifyingprocesserrors}. The integrated \ourpipeline{} pipeline significantly outperforms widely-adopted inference-time strategies, such as self-consistency~\cite{wang2023selfconsistency}, in both reasoning accuracy and token efficiency on challenging mathematical reasoning benchmarks, including AIME 2024~\cite{Aime2024}, AIME 2025~\cite{Aime2025}. Our work presents a scalable and effective approach to enhance the reliability and efficiency of complex LLM reasoning at test time, illustrated in Figure \ref{fig:combined_scaling_highlighted}. The remainder of this paper details our methodology, presents comprehensive experimental results, discusses related work, and concludes with future directions.

\section{Related Work}
\paragraph{Inference-Time Scaling Strategies.}
To navigate the inherent accuracy-efficiency trade-off in LLM reasoning, various inference-time scaling strategies increase compute at test time, such as Best-of-N sampling, self-consistency~\cite{wang2023selfconsistency}, and tree-based searches~\cite{yao2023treeofthoughts,xie2023reflexion}. While often improving accuracy, these methods can be computationally intensive and may not optimally integrate verification, sometimes exacerbating inefficiencies like ``overthinking''~\cite{chen2024overthinking}. The need for robust verification within these scaled approaches~\cite{chen2025sets,madaan2023selfrefine,gou2023critic} underscores that simply increasing generation is insufficient, calling for intelligent frameworks like \ourpipeline{} to strategically manage both generation and verification.

\begin{wrapfigure}{r}{0.35\textwidth}
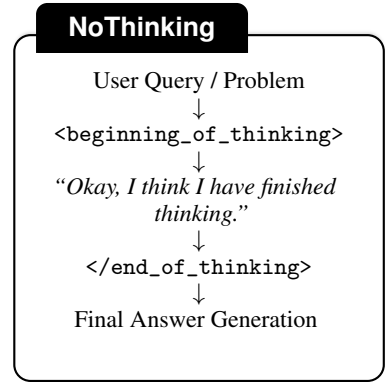
 
  \vspace{-10pt} 
  \begin{customtakeaway}[title=NoThinking] 
  \vspace{5pt}
  \centering 
  \small 
  User Query / Problem \\
  $\downarrow$ \\
  \texttt{<beginning\_of\_thinking>} \\
  $\downarrow$ \\
  \textit{``Okay, I think I have finished thinking.''} \\
  $\downarrow$ \\
  \texttt{</end\_of\_thinking>} \\
  $\downarrow$ \\
  Final Answer Generation
  \vspace{2mm} 
  \end{customtakeaway}
  \caption{The \textit{NoThinking} mechanism bypasses explicit thought generation, using a template to fill the thinking phase.}
  \label{fig:nothinking_mechanism_box}
  \vspace{-10pt} 
\end{wrapfigure}

\paragraph{Generative Process Verifiers.}
While crucial for accuracy, verifiers themselves can complicate the LLM reasoning trade-off. Expressive generative verifiers like Generative Reward Models (GenRMs) and Process Reward Models (PRMs)~\cite{lightman2023letsverify,pikwalez2024prover,saunders2022selfcritiquing,zhang2025generativeverifiersrewardmodeling,liu2025inferencetimescalinggeneralistreward} offer detailed feedback but are often computationally demanding~\cite{singhi2025solveverifycomputeoptimalproblem}, and SFT-based training may limit generalization~\cite{ouyang2022training}. Even dynamic approaches like Dyve~\cite{zhong2025dyvethinkingfastslow}, with "fast" and "slow" modes, face challenges, as per-step verification can accumulate significant overhead. In contrast, \ourmethod{}'s holistic trace analysis with dynamic budget allocation aims for a more cost-effective balance, efficiently pinpointing errors with high precision.

\paragraph{Thinking Fast and Slow in Reasoning Language Models.}
Kahneman's dual-process theory~\cite{kahneman2011thinking} informs approaches to balancing deliberate System 2-like reasoning with efficiency in LLMs~\cite{li2025system}. While some methods target generation efficiency (e.g., adaptive computation~\cite{graves2016adaptive,Diao2023BlackMamba}, pruning~\cite{zhou2023pruningcot}), extreme token reduction like the ``NoThinking'' mechanism~\cite{ma2025nothinking} highlights the \textbf{verification dilemma}: when applied to \textit{verification}, such efficiency can lead to low precision (Figure~\ref{fig:motivation}). \ourmethod{}'s dual-mode "fast" and "slow thinking" is inspired by these concepts, but its "fast thinking" is specifically optimized for \textit{reliable} error diagnosis via Reinforcement Learning~\cite{shao2024deepseekmath}. This, combined with dynamic budget allocation, seeks a more robust and efficient balance than verification strategies that are either consistently expensive or unreliably fast.



\section{Method}


\subsection{Problem Formulation}

\paragraph{System Components}
Our inference-time scaling framework uses two primary Large Language Model (LLM) components: a solver LLM and \ourmethod{}, our specialized generative verifier. Both are reasoning-capable models. The solver, an off-the-shelf LLM, generates initial candidate solutions without modification. \ourmethod{} is specifically trained for verification, with its architecture and training detailed in Section~\ref{sec:ourmethod_description}.

\paragraph{Reasoning Trace Segmentation}
A reasoning trace $S_{trace}$ is parsed into an ordered sequence of $N_s$ steps, $S_{trace} = (step_1, \ldots, step_{N_s})$. Each $step_i$ is a contiguous text segment delineated by predefined "hesitation keywords" (e.g., "hmm," as might be listed Figure~\ref{fig:hesitation_keywords_appendix_v2} in the appendix), marking transitions between keywords or the trace's start/end. This segmented trace forms the input for verification.

\begin{wrapfigure}{r}{0.6\textwidth} 
    \vspace{-10pt}
    \centering
    \includegraphics[width=\linewidth]{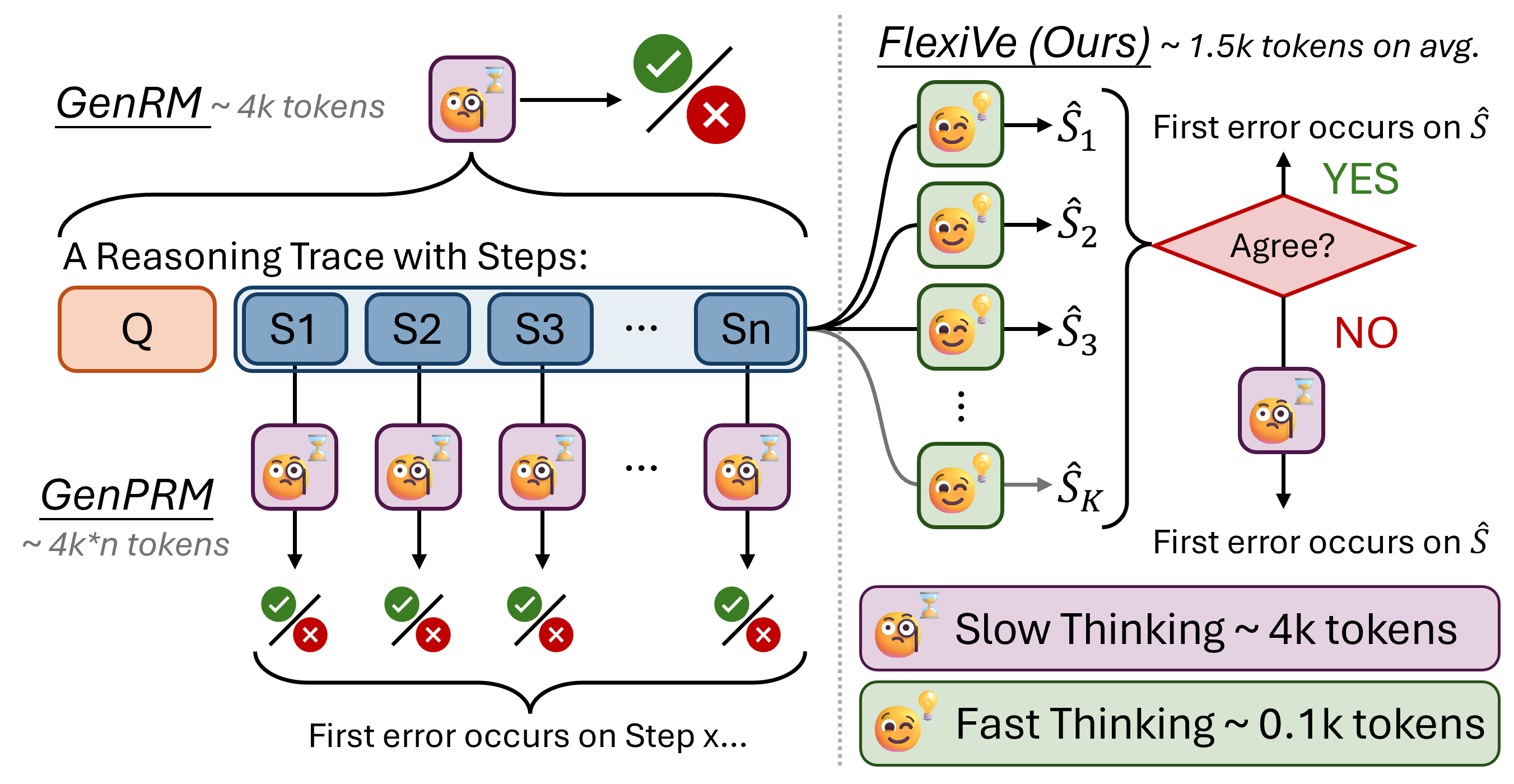} 
    \caption{Comparison of verification mechanisms. Standard GenRMs holistically assess a trace. GenPRMs often verify step-by-step. \ourmethod{} (Ours) uses an adaptive approach on the entire trace, with initial parallel fast evaluations deciding if deeper, slow verification is needed.}
    \label{fig:compare_mechanisms_wrapped}
    \vspace{-5pt} 
\end{wrapfigure}

\paragraph{Verifier Operation and Output}
The task of the verifier is to assess the correctness of the solver's reasoning trace $S_{trace}$. Different verifier architectures approach this differently, as illustrated in Figure~\ref{fig:compare_mechanisms_wrapped}. For example, a standard Generative Reward Model (GenRM) might perform a single, comprehensive "long thinking" pass over the entire query and trace to output a binary judgment. Process-focused variants like GenPRM often conduct sequential, step-by-step verification, which can be computationally intensive.
In our framework, the steps from $S_{trace}$ are formatted using a critic template~\cite{zheng2024processbenchidentifyingprocesserrors} to create an input prompt for \ourmethod{}. Unlike per-step verifiers, \ourmethod{} evaluates the entire trace but employs a dynamic strategy (detailed in Section~\ref{sec:ourmethod_description}) to modulate its computational effort. It outputs $V_{out} = (F, idx_{pred})$, where $F$ is a textual error analysis and $idx_{pred}$ is the predicted index of the first error. Consistent with its training, $idx_{pred} = -1$ signifies no errors. This $V_{out}$ informs decisions within our \ourpipeline{}.

\subsection{\ourmethod{}}
\label{sec:ourmethod_description} 
\ourmethod{} is a generative verifier that dynamically modulates computational effort during test-time verification, operating in "fast thinking" and "slow thinking" modes. The fast thinking mode, inspired by~\citet{ma2025nothinking} and enhanced with Reinforcement Finetuning, generates significantly shorter outputs 27x-40x in Figure ~\ref{fig:motivation}) than the conventional slow thinking mode's detailed trace. Our Flexible Allocation of Verification Budget scheme manages these modes to leverage fast thinking's efficiency.
\paragraph{Reinforcement Training}
\ourmethod{} is trained using Group Relative Policy Optimization (GRPO) \cite{shao2024deepseekmath}. In this framework, a base reasoning model fine-tuned on a mistake detection task predicts either the index of the first error (\(idx_{gt}\)) or returns \(-1\) if the reasoning is correct. GRPO optimizes the model’s generation policy by maximizing a composite reward defined as \(R_i = R_{\text{correct}} + R_{\text{length}}\).

The correctness reward $R_{\text{correct}}$ is defined by
\begin{equation}
    R_{\text{correct}}(idx_{pred}, idx_{gt}) = \begin{cases} 1.0 & \text{if } idx_{pred} = idx_{gt} \\ 0.0 & \text{otherwise} \end{cases},
\end{equation}
assigning a binary score based on the match between the predicted and true error indices.


The length adjustment reward $R_{\text{length}}$ modulates the response length \(L\) relative to \(idx_{gt}\) and is given by $R_{\text{length}} = -P(L, idx_{gt})$, where \(P(L, idx_{gt})\) is the length penalty function. 

\begin{equation}
  P(L, idx_{gt}) = 
  \begin{cases}
    \min\left(P_{max}, c_{fast} \cdot \max(0, L - L_{fast})\right) & \text{if } idx_{gt} = -1 \\[1ex]
    \begin{alignedat}{1}
        &\min\left(P_{max}, c_{under} \cdot \max(0, L_{slow\_min} - L)\right) \\
      {}+{}&\min\left(P_{max}, c_{over} \cdot \max(0, L - L_{slow\_max})\right)
    \end{alignedat} & \text{if } idx_{gt} \neq -1
  \end{cases}
\label{eq:penalty_reward} 
\end{equation}

In Equation~\ref{eq:penalty_reward}, when \( idx_{gt} = -1 \), responses exceeding the target length \( L_{fast} \) are penalized, thereby promoting “fast thinking” in the absence of errors. Conversely, if \( idx_{gt} \neq -1 \), lengths outside the interval \([L_{slow\_min}, L_{slow\_max}]\) are penalized to encourage “detailed thinking” during error analysis. Training involves sampling \( G \) outputs per prompt, computing each reward, and calculating advantages relative to the group's average as in ~\citet{shao2024deepseekmath}.

\paragraph{Flexible Allocation of Verification Budget}
\ourmethod{} dynamically allocates its verification budget. The core intuition is thus to leverage inexpensive, parallelizable probes to gauge verification difficulty upfront, and only escalate to more resource-intensive analysis when these probes indicate ambiguity or complexity, thereby tailoring computational effort to the specific needs of each verification instance. Initially, it performs \( k \) "fast thinking" verification runs. The consensus among these runs is measured by the agreement ratio:
\begin{equation}
R_{\text{agreement}} = \frac{\max_{i} a_i}{k},
\end{equation}
where \( a_i \) is the count of the most frequent outcome (using fuzzy error index matching). If this ratio meets a predefined threshold \( \tau \) (\( R_{\text{agreement}} \geq \tau \)), the consensus result from the fast phase, \( V_{\text{fast}} \), is accepted. Otherwise, \( \max(1, k/8) \) additional, resource-intensive "slow thinking" runs are triggered to produce the final outcome \( V_{\text{slow}} \). The overall verification result \( V \) is thus determined by:
\begin{equation}
V =
\begin{cases}
V_{\text{fast}}, & \text{if } R_{\text{agreement}} \geq \tau, \\
V_{\text{slow}}, & \text{otherwise}.
\end{cases}
\end{equation}

\begin{wrapfigure}{r}{0.55\textwidth} 
  \vspace{-20pt} 
  \begin{minipage}{\linewidth} 
    \begin{algorithm}[H] 
      \caption{Solve-Detect Stage of \ourpipeline{}}
      \label{alg:solve_detect_wrapped}
      \fontsize{7.5pt}{9.5pt}\selectfont 
      \begin{algorithmic}[1]
        \Statex \textbf{Input:} Problem $P$, Solver $\Msolve$
        \Statex \textbf{Output:} Candidate Solution $S_1$
        \Procedure{SolveDetect}{$P, \Msolve$}
          \State $S_1 \leftarrow \emptyset$
          \State $stop\_flag \leftarrow \text{false}$
          \For{$k = 1$ \textbf{to} $L_{max}$} \Comment{$L_{max}$ is max length}
            \State $t_k \sim \Msolve(\cdot | P, S_1^{(k-1)})$
            \State $S_1^{(k)} \leftarrow S_1^{(k-1)} \oplus t_k$
            \If{$t_k = \EOS$}
                \State $stop\_flag \leftarrow \text{true}$
            \EndIf
            \If{$S_1^{(k)}$ ends with $kw \in \Khesitation$}
                \State $logp_{\text{Yes}} \leftarrow -\log p_{\Msolve}(\text{Yes} | \PromptComplete{S_1^{(k)}})$
                \State $logp_{\text{No}} \leftarrow -\log p_{\Msolve}(\text{No} | \PromptComplete{S_1^{(k)}})$
                \If{$logp_{\text{Yes}} > logp_{\text{No}}$} \Comment{Compare log-probs}
                     \State $stop\_flag \leftarrow \text{true}$ \Comment{Solution complete}
                \EndIf
            \EndIf
            \If{$stop\_flag$}
                \State \textbf{break}
            \EndIf
            \State $S_1 \leftarrow S_1^{(k)}$
          \EndFor
          \State \textbf{return} $S_1$
        \EndProcedure
      \end{algorithmic}
    \end{algorithm}
  \end{minipage}
  \vspace{-20pt} 
\end{wrapfigure}

This adaptive strategy optimizes computational cost by reserving intensive verification only for cases where initial fast assessments lack sufficient agreement. Crucially, this decision logic and the subsequent verification (whether fast or slow) are applied to the reasoning trace as a whole, rather than on a per-step basis as in some prior dynamic verifiers like Dyve~\cite{zhong2025dyvethinkingfastslow}. By evaluating the entire trace with a dynamically chosen verification depth, \ourmethod{} aims to avoid the accumulated cost of per-step decisions, potentially offering better scalability and efficiency, especially for longer or more complex reasoning processes. The intuition is to use a quick, broad assessment first, and only invest significant resources when this initial assessment signals higher uncertainty or difficulty.

\subsection{\ourpipeline{}}
\ourpipeline{} is a multi-component framework designed to enhance the reasoning accuracy and efficiency of Large Language Models (LLMs). The pipeline integrates distinct modules: an initial solution generation phase (Solve), a mid-stream reasoning monitoring and management stage (Detect), and a combined validation and conditional refinement process (Verify and Refine). The complete pipeline is summarized in Algorithm~\ref{apdx:implmentation_pipeline_appendix}, with detailed implementation provided in Appendix. The conceptual framework of the pipeline is as follows:

\paragraph{Solve}
The `Solve' stage initiates the process, wherein the solver LLM is tasked with generating an initial, step-by-step candidate solution ($S_1$) to a given problem. This stage forms the foundational attempt at problem-solving, producing a complete reasoning trace and a final answer for subsequent evaluation.

\paragraph{Detect}
The `Detect' module continuously monitors LLM output for predefined hesitation keywords (Figure~\ref{fig:hesitation_keywords_appendix_v2} in the Appendix). Upon detecting a keyword, generation pauses, and the LLM is prompted (Figure~\ref{fig:detect-prompt_appendix_v2} in the Appendix) to assess solution completeness via log-probabilities (\(-\log p(\text{Yes})\) vs. \(-\log p(\text{No})\)). This check efficiently reuses over 90\% of the generation prefix, preserving the Key-Value (KV) cache and minimizing computation overheads. If reasoning is deemed complete, the pipeline advances to `Verify and Refine'; otherwise, generation resumes. This adaptive monitoring reduces overhead and enables early verification.

\paragraph{Verify and Refine}
Upon full generation or early completion detected by the `Detect' module, the candidate solution $S_1$ is assessed by \ourmethod{}, which identifies any errors and their specific step $idx_{pred}$. A validated $S_1$ directly becomes the final output. If an error is found in $S_1$, \ourmethod{}'s diagnostic feedback ($F_1$) guides the solver LLM to generate a single new candidate solution, $S_2$, aiming to correct the error by exploring an alternative reasoning path. This refined solution $S_2$ is then accepted as the final output, without requiring an additional validation round. This integrated approach of validation followed by conditional, feedback-driven refinement ensures a balance between rigorous solution assessment and efficient improvement.

\section{Experiments}
\label{sec:experiments}

Our experiments are designed to achieve two primary goals. First, we evaluate the performance and efficiency of \ourmethod{} as a standalone generative verifier, analyzing its scaling properties compared to baseline approaches. Second, we assess the effectiveness of our \ourpipeline{} in enhancing the reasoning accuracy and computational efficiency of LLMs on complex mathematical tasks, comparing it against standard inference-time strategies.



\subsection{Experimental Setup}
\label{sec:exp_setup}

For detailed experimental configurations, including hyperparameter settings for all models and full dataset statistics, please refer to Appendix ~\ref{apdx:exp_setup_appendix}.

\paragraph{\ourmethod{} Training} \ourmethod{} is initialized from DeepSeek-R1-Distill-Qwen-14B~\cite{shao2024deepseekmath} and trained for mistake detection using Group Relative Policy Optimization (GRPO)~\cite{shao2024deepseekmath} on 90\% of the BIG-Bench Mistake dataset~\cite{tyen-etal-2024-llms}, with 10\% for validation. All the NoThinking mechanism are activated for all input problem and reasoning traces pair to ensure that the training was performed in `fast mode' for targeted improvement. The objective, optimizing a composite reward, uses LORA PEFT~\cite{hu2021lora} ($r=16, \alpha=32$) and AdamW~\cite{loshchilov2017decoupled}. Key GRPO parameters include $G=14$ samples per input and a KL coefficient of 0.04. All experiments are conducted on 8 $\times$ NVIDIA A800-SXM4-80GB GPUs.


\paragraph{Evaluation Tasks and Datasets}
We assess \ourmethod{}'s step-level verification capability, measured by F1 score, on the comprehensive ProcessBench benchmark~\cite{zheng2024processbenchidentifyingprocesserrors}. ProcessBench includes diverse mathematical reasoning datasets such as GSM8K, MATH, OlympiadBench, and OmniMATH. For the full \ourpipeline{}, we evaluate end-to-end task accuracy and token efficiency on particularly challenging mathematical datasets: AIME (2024, 2025)~\cite{Aime2024, Aime2025}, AMC, CNMO~\cite{liu2024your}, and OlympiadBench. Especially, All token counts in the results refer exclusively to the output tokens generated by the LLM, in the entire testing dataset.

\paragraph{Baselines}
On ProcessBench, \ourmethod{}'s performance is compared against established Process Reward Models (PRMs) from~\cite{zheng2024processbenchidentifyingprocesserrors}, as detailed in Table~\ref{tab:processbench_results}. We also include a comparison with a token-efficient `NoThinking' verification approach, similar to that described in~\cite{ma2025nothinking}, which represents a simpler, non-deliberative verification strategy.
For evaluating the \ourpipeline{}, DeepSeek-R1 14B and 32B models~\cite{shao2024deepseekmath} serve as the base "worker" LLMs. The pipeline's performance is benchmarked against: (1) the direct output of the worker LLM, representing a standard prompting baseline, and (2) Self-Consistency with majority voting~\cite{wang2023selfconsistency}, a widely recognized inference-time technique for enhancing LLM reasoning by sampling multiple solutions.

\begin{table*}[hbt!]
\centering
\caption{ProcessBench results reported with F1 scores. Results for \textbf{\ourmethod} are highlighted . \textbf{bold} indicates the best in the sub category. All \ourmethod~variants are trained on only 1526 samples.}
\resizebox{1.0\textwidth}{!}{
\begin{tabular}{lcccccc}
\toprule
\multirow{2}{*}{\textbf{Model}} & \multirow{2}{*}{\textbf{\# Samples}} & \multirow{2}{*}{\textbf{GSM8K}} & \multirow{2}{*}{\textbf{MATH}} & \multirow{2}{*}{\begin{tabular}[c]{@{}c@{}} \bf Olympiad \\ \bf Bench \end{tabular}} & \multirow{2}{*}{\begin{tabular}[c]{@{}c@{}} \bf Omni- \\ \bf MATH \end{tabular}} & \multirow{2}{*}{\textbf{Avg.}} \\
& \\
\midrule
\multicolumn{7}{c}{\textit{Proprietary Models}} \\
\midrule
GPT-4o-0806                      & unk          & 79.2          & 63.6          & 51.4          & 53.5          & 61.9 \\
o1-mini                           & unk          & 93.2          & 88.9          & 87.2          & 82.4          & 87.9 \\
\midrule
\multicolumn{7}{c}{\textit{Open Source Models (7-8B)}} \\
\midrule
Qwen2.5-Math-PRM-7B               & $\sim$344K   & 82.4          & 77.6          & 67.5          & 66.3          & 73.5 \\
RetrievalPRM-7B                   & 404K         & 74.6          & 71.1          & 60.2          & 57.3          & 65.8 \\
Universal-PRM-7B                  & unk          & 85.8          & 77.7          & 67.6          & 66.4          & 74.3 \\
Direct Generative PRM-7B          & 23K          & 63.9          & 65.8          & 54.5          & 55.9          & 60.0 \\
\rowcolor{gray!10} \textcolor{gray}{GenPRM-7B w/ Code Exec (Pass@1)}        & \textcolor{gray}{23K}          & \textcolor{gray}{78.7}          & \textcolor{gray}{80.3}          & \textcolor{gray}{72.2}          & \textcolor{gray}{69.8}          & \textcolor{gray}{75.2} \\
\rowcolor{gray!10} \textcolor{gray}{GenPRM-7B w/ Code Exec (Maj@8)}         & \textcolor{gray}{23K}          & \textcolor{gray}{81.0}          & \textcolor{gray}{85.7}          & \textcolor{gray}{78.4}          & \textcolor{gray}{76.8}          & \textcolor{gray}{80.5} \\
\midrule
\multicolumn{7}{c}{\textit{Open Source Models (14-32B) w/ \textbf{Moderate Compute}}} \\
\midrule
Dyve-14B                          & 117K         & 68.5          & 58.3          & 49.0          & 47.2          & 55.8 \\
GenPRM-32B w/o Code Exec (Maj@8)  & 23K          & 78.8          & \uline{85.1}    & 78.7          & \uline{74.9}    & 79.3 \\
\rowcolor{violet!10}\ourmethod (Flex@32)                     & \textbf{1526}         & \uline{82.8}    & 83.3          & \uline{79.2}    & 73.4          & \uline{79.7} \\
\rowcolor{violet!10}\ourmethod (Flex@128)                    & \textbf{1526}         & \textbf{83.0} & \textbf{85.0} & \textbf{80.0} & \textbf{75.2} & \textbf{80.8} \\
\midrule
\multicolumn{7}{c}{\textit{Open Source Models (14-32B) w/ \textbf{High Compute}}} \\
\midrule
\rowcolor{gray!10} GenPRM-32B (Pass@1) w/ Code Exec      & 23K          & 83.1          & 81.7          & 72.8          & 72.8          & 77.6 \\
\rowcolor{gray!10} GenPRM-32B (Maj@8) w/ Code Exec       & 23K          & 85.1          & 86.3          & 78.9          & 80.1          & 82.6 \\
\rowcolor{violet!10}\ourmethod (Think@64)                      & \textbf{1526}         & \textbf{88.1} & \textbf{90.1}  & \textbf{86.7}   & \textbf{80.4}             & \textbf{86.3}    \\
\bottomrule
\end{tabular}
}
\label{tab:processbench_results}%
\end{table*}%


\subsection{\ourmethod{} Performance and Scaling Analysis}
\label{sec:ourmethod_performance}

This section evaluates \ourmethod{}'s error identification accuracy on ProcessBench~\cite{zheng2024processbenchidentifyingprocesserrors} and its efficiency on subsets GSM8K and MATH. We test \ourmethod{} in several configurations: \textbf{\ourmethod{} (Flex@k)} uses adaptive verification, starting with $k$ initial "fast" verification samples and dynamically deciding whether to escalate to more thorough verification; \textbf{\ourmethod{} (Think@k)} employs $k$ samples from \ourmethod{}'s deliberative "slow thinking" verification process with majority vote, designed for higher accuracy at a typically higher computational cost; and \textbf{\ourmethod{} (NoThinking@k)} represents \ourmethod{} in a purely "fast thinking" or non-deliberative mode using $k$ samples with majority vote, analogous to the `NoThinking' baseline but with \ourmethod{}'s architecture. The "Moderate Compute" and "High Compute" categories in Table~\ref{tab:processbench_results} are broadly defined by the number of verification samples or overall inference cost, with "High Compute" settings involving more extensive verification efforts.

\begin{wrapfigure}{r}{0.65\columnwidth}
    \centering
    \includegraphics[width=\linewidth]{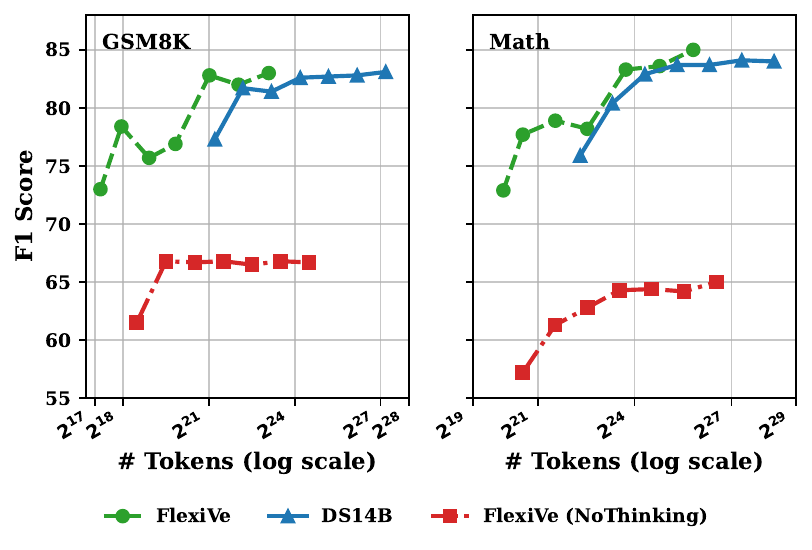}
    \vspace{-20pt}
    \caption{F1 score vs. verification tokens on GSM8K (left) and MATH (right). \ourmethod (Flex@k, green circles) demonstrates higher F1 for similar token costs than \texttt{DeepSeek-R1-Distill-Qwen-14B} (blue triangles, baseline verifier), both outperforming the token-efficient \ourmethod (NoThinking variant, red squares). X-axis denote the number of token generated across the entire test set. }
    \label{fig:f1_vs_tokens}
    \vspace{-10pt}
\end{wrapfigure}

\paragraph{Verification Accuracy on ProcessBench}
Table~\ref{tab:processbench_results} (with \ourmethod{} results highlighted in violet 
) details the F1 scores for \ourmethod{} compared to various baselines.
In the "Moderate Compute" setting, \ourmethod{} (Flex@128) achieves a strong average F1 score of 80.8\%, with a notable 85.0\% on the MATH dataset. This performance surpasses the GenPRM-32B (Maj@8) model (without code execution), which scores 79.3\% average F1, despite \ourmethod{} being trained on significantly fewer samples (1,526 vs. 23K). The \ourmethod{} (Flex@32) configuration also demonstrates competitive performance with a 79.7\% average F1 score.

In the "High Compute" setting, \ourmethod{} (Think@64), utilizing its deliberative "slow thinking" mode, achieves exceptional F1 scores of 88.1\% on GSM8K and 90.1\% on MATH. This performance notably exceeds that of the compute-intensive GenPRM-32B (Maj@8) model with code execution (which scores 85.1\% on GSM8K and 86.3\% on MATH). These results highlight that \ourmethod{}'s sophisticated deliberative verification (\texttt{Think@64}), despite its own computational demands, can achieve superior accuracy compared to other large verifiers, even those augmented with code execution. This underscores the effectiveness of \ourmethod{}'s architecture and training, even when scaled to more intensive verification tasks. The significantly smaller training data requirement for \ourmethod{} across all its configurations further emphasizes its sample efficiency.

\paragraph{Efficiency and Budget Scaling (GSM8K \& MATH)}
Figure~\ref{fig:f1_vs_tokens} depicts the accuracy-cost trade-off. On both GSM8K and MATH, \ourmethod{} (Flex@k) (green circles) provides a better F1 score for comparable token usage than the baseline verifier, \texttt{DeepSeek-R1-Distill-Qwen-14B} (DS14B, blue triangles). While both \ourmethod{} (Flex@k) and DS14B reach higher peak F1 scores, the \ourmethod{} (NoThinking@k) variant (red squares) is considerably more token-frugal, albeit with a lower F1 ceiling.

\subsection{Scaling \ourpipeline{} for Enhanced Performance}
\label{sec:ourpipeline_scaling}

We evaluate \ourpipeline{} on AIME2024~\cite{Aime2024}, AIME2025~\cite{Aime2025}, and CNMO~\cite{liu2024your} to understand its scaling properties. We explore two primary scaling dimensions: first, varying \ourmethod{}'s verification budget within a single pipeline execution, and second, generating multiple candidate solutions from the worker LLM, each processed by \ourpipeline{}.

\begin{wrapfigure}{r}{0.65\columnwidth}
    \centering
    \vspace{-5pt} 
    \includegraphics[width=\linewidth]{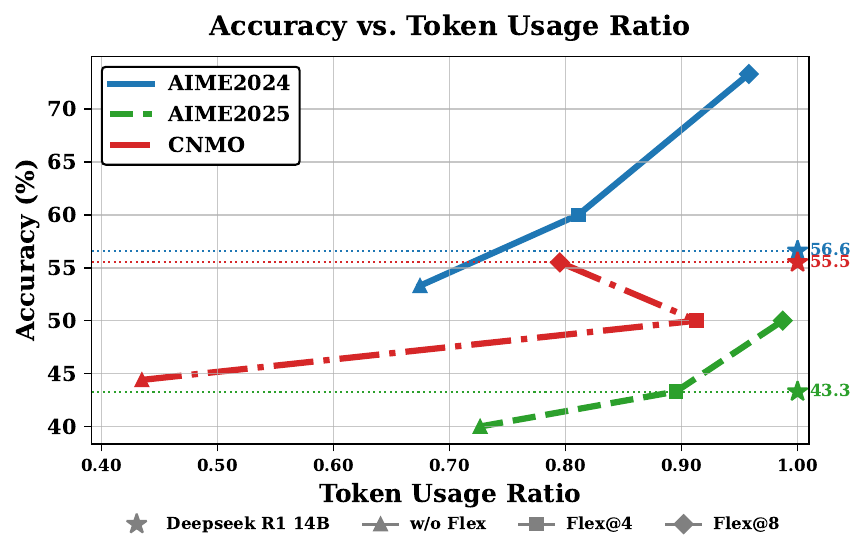} 
    \caption{Impact of scaling \ourmethod{}'s verification budget (Flex@N) within a single \ourpipeline{} execution on Pass@1 Accuracy vs. Token Usage Ratio relative to DeepSeek R1 14B. Benchmarks are color/linestyle distinguished. }
    \label{fig:scaling_fast_thinking} 
    \vspace{-5pt} 
\end{wrapfigure}

\paragraph{Scaling \ourmethod{} Verification Budget (Flex@N) in a Single Pipeline Run}
We first analyze scaling \ourmethod{}'s internal verification budget (`Flex@N`, representing N fast-thinking verification samples post-extraction) within a single pipeline pass. In figure~\ref{fig:scaling_fast_thinking}, the `w/o Flex' setup (`Solve + Detect') significantly cuts token usage, token ratio ~0.67 on AIME2024 and ~0.43 on CNMO, but can reduce accuracy, notably on CNMO (44.4\% vs. 55.5\% baseline). Integrating \ourmethod{} verification, particularly `Flex@8`, substantially boosts accuracy over baseline on AIME2024 (73.3\% vs. 56.6\%) and AIME2025 (50.0\% vs. 43.3\%), and matches baseline accuracy on CNMO (55.5\%). Crucially, these `Flex@8` configurations use fewer tokens than the baseline (e.g., ~0.96 AIME2024, ~0.80 CNMO token ratio), demonstrating \ourpipeline{}'s token-efficient accuracy gains. However, CNMO's less consistent improvement with N suggests that varying only the verifier budget might not universally ensure peak performance.

\paragraph{Scaling Solver and Verifier via Multiple Solutions}
To achieve more consistent gains and higher peak accuracies, we scale compute by generating multiple solutions from the solver, each verified by \ourmethod{}. On the AIME2024 benchmark (Figure~\ref{fig:combined_scaling_highlighted}, left panel), this strategy yields significant and consistent accuracy improvements as more solutions are processed: accuracy climbs from ~67.5\% (1 solution) to over 83\% (16 solutions). This approach effectively leverages increased solver compute, with \ourmethod{} identifying the correct solution among candidates, demonstrating a robust path to superior performance, especially for top-tier accuracy. 
This underscores our takeaway in Figure ~\ref{fig:takeaway}: for optimal results with \ourpipeline{}, scaling solver LLM's compute is as important as scaling \ourmethod{}'s verification capabilities.


\begin{figure}[hb]
\vspace{-10pt}
\begin{customtakeaway}[title=Takeaway for \ourpipeline scaling] 
  \vspace{5pt}
  With \ourpipeline, scaling solver LLM's compute is as important as scaling \ourmethod{}'s.
\end{customtakeaway}
\vspace{-5pt}
\caption{A take away highlights the symbiotic relationship.}
\vspace{-5pt}
\label{fig:takeaway}
\end{figure}

\subsection{Extended Analysis}
\label{subsec:extended_analysis}

\begin{wrapfigure}{r}{0.4\columnwidth} 
    \centering
    \vspace{-5pt} 
    \includegraphics[width=\linewidth]{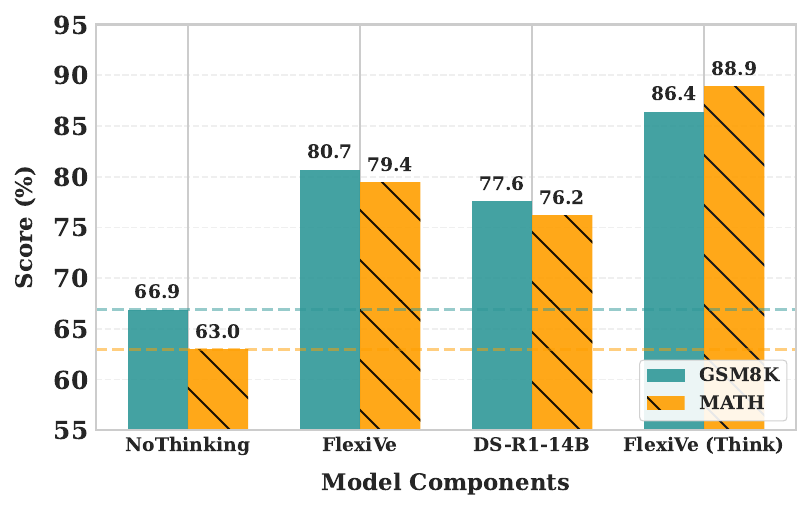}
    \caption{Ablation: Component impact on GSM8K/MATH (\%). \ourmethod (Think) excels; \ourmethod (Flex@4) also surpasses NoThinking (maj@8) and DS-R1-14B (Think@1). }
    \label{fig:ablation_components}
    \vspace{-20pt} 
\end{wrapfigure}

\paragraph{Component Performance Comparison}
\label{par:component_comparison}
An ablation study assessed individual component impacts. For \ourmethod, we used Flex@4; for NoThinking, maj@8; and for both the \texttt{DeepSeek-R1-Distill-Qwen-14B} baseline and \ourmethod's deliberative mode, Think@1, ensuring roughly comparable computational budgets.
Figure \ref{fig:ablation_components} shows that \ourmethod's Reinforcement Learning (RL) training not only matches or slightly exceeds the baseline verifier's performance under similar compute but also significantly outperforms when \ourmethod engages its "thinking" mode. This is crucial: though trained with RL primarily leveraging its efficient "NoThinking" (fast) mode, \ourmethod generalizes effectively to improve verification in its more deliberative "thinking" mode, underscoring its RL-trained robustness and adaptability.

\begin{wrapfigure}{r}{0.4\columnwidth} 
    \centering
    \vspace{-10pt} 
    \includegraphics[width=\linewidth]{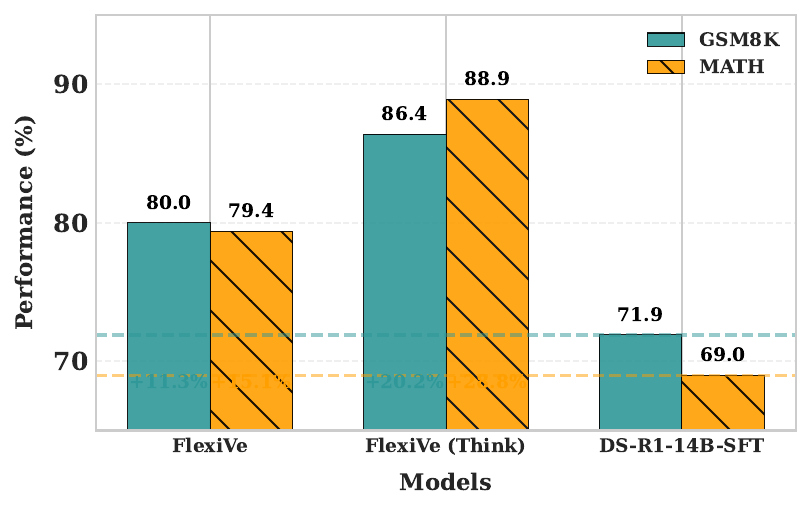}
    \caption{RL (\ourmethod) vs. SFT ({DeepSeek-R1-Distill-Qwen-14B}) on GSM8K/MATH. RL-trained \ourmethod, especially in thinking mode, shows superior generalization over the SFT baseline.}
    \label{fig:rl_vs_sft}
    \vspace{-25pt} 
\end{wrapfigure}

\paragraph{RL vs. SFT}
\label{par:rl_vs_sft}
We compared our RL approach with traditional SFT for training verifiers. The SFT baseline used 10,000 reasoning paths with problems form OpenO1~\cite{openo1_sft_ultra_dataset} and generated by \texttt{DeepSeek-R1-Distill-Qwen-14B}. They are labeled via LLM-based judging as ~\cite{zheng2024processbenchidentifyingprocesserrors}.
Findings (Figure \ref{fig:rl_vs_sft}) suggest SFT lack generalization. Reasoning traces in benchmarks like ProcessBench, often from weaker, non-thinking LLMs, are shorter and less complex. This led to performance drops for SFT verifier on more diverse processes. In contrast, \ourmethod, RL-trained on only 1,526 BIG-Bench Mistake~\cite{tyen-etal-2024-llms} problems, showed strong generalization. This highlights RL's advantage in fostering robust verifiers with significantly less data than typical SFT.


\section{Limitations}
\label{sec:limitations}

While \ourmethod{} and \ourpipeline{} demonstrate promising advancements, several avenues warrant future investigation to enhance their robustness and broaden their applicability. The generalization of \ourmethod{} is inherently linked to its training data diversity, and our current validation, primarily on mathematical reasoning due to computational constraints, invites further cross-domain exploration (e.g., in program synthesis or commonsense QA). The empirically-set parameters ($k, \tau$) for \ourmethod{}'s dynamic budget allocation would benefit from a comprehensive sensitivity analysis and the development of automated tuning guidelines to maximize practical adoption. Furthermore, although \ourpipeline{} is designed for efficiency—with mechanisms like KV cache reuse in its heuristic `Detect' stage—its multi-component nature and dynamic mode-switching introduce inherent computational overhead. We believe this overhead could be substantially mitigated, and overall performance significantly boosted, through optimized implementations, potentially leveraging advanced inference engines like vLLM~\cite{kwon2023efficient} or SGLang~\cite{zheng2024sglang}; advancing this represents a valuable direction for community exploration to fully realize the benefits of such dynamic reasoning systems. Addressing these aspects will be key to the continued development and deployment of sophisticated, efficient, and widely applicable verified reasoning frameworks.

\section{Conclusion}
\label{sec:conclusion}

We introduce \ourmethod{}, a dynamic verifier balancing computational cost and accuracy, integrated into the \ourpipeline{} pipeline for efficient LLM reasoning enhancement. Experiments confirm that our pipeline, leveraging \ourmethod{}, achieves significant gains in both accuracy and token efficiency over baselines, highlighting flexible verification and intelligent pipeline design as a scalable path toward more reliable and efficient complex reasoning in LLMs.

\newpage

\newpage
\bibliographystyle{unsrtnat}
\bibliography{References}

\newpage
\appendix

\section{Appendix}

\subsection{Extended Experimental Setup}
\label{apdx:exp_setup_appendix}

\paragraph{\texttt{FlexiVe} Training.}
\label{par:flexive_training_appendix}
We train \texttt{FlexiVe} using Group Relative Policy Optimization (GRPO) \cite{shao2024deepseekmath} on a mistake detection task. The policy $\pi_{\theta}$ is initialized from the DeepSeek-R1-Distill-Qwen-14B model \cite{shao2024deepseekmath}. 
We utilize the BIG-Bench Mistake dataset \cite{tyen-etal-2024-llms}, reserving 90\% for training and 10\% for validation. The training objective is to predict the index of the first reasoning error ($idx_{gt}$) or output -1 if the trace is correct, optimized using the composite reward detailed in Section \ref{sec:experiments} (main paper). 
Parameter-Efficient Fine-Tuning (PEFT) is employed via LoRA \cite{hu2021lora} with rank $r=16$ and $\alpha=32$, targeting the attention projection layers. Optimization is performed using AdamW \cite{loshchilov2017decoupled} with a learning rate of $5 \times 10^{-6}$ and gradient checkpointing. For GRPO, we sample $G=14$ outputs per input, and the KL coefficient is set to 0.04. Training is managed using the \texttt{transformers} \cite{wolf-etal-2020-transformers} and \texttt{trl} \cite{trl_library} libraries, with experiment tracking via Weights \& Biases \cite{wandb}.

\paragraph{Evaluation Tasks and Datasets.}
\label{par:eval_datasets_appendix}
For \textbf{\texttt{FlexiVe} evaluation}, to assess its step-level verification capabilities, we use the ProcessBench benchmark \cite{zheng2024processbenchidentifyingprocesserrors}. This includes diverse mathematical reasoning datasets such as GSM8K, MATH, OlympiadBench, and OmniMATH. Performance is measured using the F1 score for identifying the first erroneous step.
For \textbf{\texttt{Solve-Detect-Verify} pipeline evaluation}, to assess end-to-end effectiveness, we use a suite of challenging mathematical reasoning datasets: AIME (2024 and 2025) \cite{Aime2024, Aime2025}, AMC (mentioned in main text, details can be added if necessary), CNMO \cite{liu2024your}, and OlympiadBench (also used for \texttt{FlexiVe} evaluation). AIME is a prestigious high school mathematics competition known for its challenging mathematical problems, and contains problems from the American Invitational Mathematics Examination (AIME) 2024 and 2025. The CNMO Benchmark evaluates AI on China’s National Mathematical Olympiad problems, focusing on advanced proof-based problem-solving.
On these tasks, we measure final task accuracy and computational efficiency (e.g., total tokens).

\paragraph{Baselines.}
\label{par:baselines_appendix}
For \textbf{\texttt{FlexiVe} baselines} on ProcessBench, we compare against state-of-the-art Process Reward Models (PRMs) as reported in \cite{zheng2024processbenchidentifyingprocesserrors} and the 'NoThinking' verification approach adapted from \cite{ma2025reasoning} (or your specific citation, e.g., \cite{ma2025nothinking}).
For our \textbf{\texttt{Solve-Detect-Verify} pipeline}, we use DeepSeek-R1 14B \cite{shao2024deepseekmath} as the base worker LLMs. The full pipeline is compared against: (1) the worker LM generating solutions directly (potentially with the 'Detect' mechanism only), and (2) the Self-Consistency method \cite{wang2023selfconsistency} applied to the worker LM.

\subsection{\texttt{Solve-Detect-Verify} Pipeline Implementation Details}
\label{apdx:implmentation_pipeline_appendix}

The \texttt{Solve-Detect-Verify} pipeline is implemented with a specific two-attempt strategy derived from our Python codebase, emphasizing adaptive verification and intelligent solution generation. Algorithm~\ref{alg:sdv_pipeline_python_logic_appendix} outlines this refined flow. Key components like solution generation with hesitation detection and adaptive verification are encapsulated in helper functions for clarity.

\begin{algorithm}[H]
\caption{\texttt{Solve-Detect-Verify} Pipeline (Reflecting Python Implementation Logic)}
\label{alg:sdv_pipeline_python_logic_appendix}
\begin{algorithmic}[1]
\Require Problem $P$, Verification Parameters $\Theta_V = (k_{fast}, \tau_{agree}, k_{slow})$, Best-of-N $N_{BoN}$
\State $S_{final} \gets \text{NIL}$
\State \Comment{--- Attempt 1: Initial Solve and Adaptive Verification ---}
\State $Prompt_1 \gets \text{FormatInitialPrompt}(P)$
\State $S_1 \gets \text{GenerateSolutionWithDetection}(LLM, Prompt_1)$ \Comment{Handles streaming, hesitation detection, and continuation}
\State $(\text{is\_valid}_1, \text{error\_step}_1, F_1) \gets \text{AdaptiveVerify}(P, S_1, \Theta_V)$ \Comment{Uses $k_{fast}, \tau_{agree}, k_{slow}$}
\If{$\text{is\_valid}_1 = \text{True}$}
    \State $S_{final} \gets S_1$
\Else \Comment{Attempt 1 failed or was deemed invalid by verification}
    \State \Comment{--- Attempt 2: Retry with Best-of-N (BoN) ---}
    \State $Prompt_2 \gets \text{FormatRetryPromptWithFeedback}(P, S_1, F_1)$
    \State $Solutions_{candidates} \gets []$; $Answers_{candidates} \gets []$
    \For{$i = 1$ \textbf{to} $N_{BoN}$}
        \State $S_{cand} \gets \text{GenerateSolutionWithDetection}(LLM, Prompt_2)$
        \State Add $S_{cand}$ to $Solutions_{candidates}$
        \State $Ans_{cand} \gets \text{ExtractFinalAnswer}(S_{cand})$
        \State Add $Ans_{cand}$ to $Answers_{candidates}$
    \EndFor
    \State $(Ans_{majority}, S_{voted}) \gets \text{MajorityVote}(Answers_{candidates}, Solutions_{candidates})$
    \State $S_{final} \gets S_{voted}$ \Comment{BoN result is used directly without re-verification as per Python code}
\EndIf
\If{$S_{final}$ is NIL and $S_1$ is not NIL} \Comment{Fallback if BoN stage wasn't reached or produced nothing, but $S_1$ exists}
    \State $S_{final} \gets S_1$ 
\EndIf
\State Evaluate $S_{final}$ against ground truth.
\State \Return $S_{final}$, Evaluation Metrics, Total Compute Cost
\end{algorithmic}
\end{algorithm}

\paragraph{Key Helper Functions:}
\begin{itemize}
    \item \textbf{\texttt{GenerateSolutionWithDetection}(LLM, Prompt)}: This function generates a solution by streaming tokens from the LLM. It incorporates the hesitation detection mechanism (detailed below in the "Detect" section) to identify potential points of self-correction or solution completion. If hesitation is detected and the solution is deemed complete by an internal check, generation might be paused and then explicitly continued to ensure the full thought process is captured before final truncation.
    \item \textbf{\texttt{AdaptiveVerify}(P, S, $\Theta_V$)}: This function performs verification on solution $S$ for problem $P$. It first conducts $k_{fast}$ "fast thinking" verifications. If the agreement ratio among these (based on exact error step matching) meets or exceeds $\tau_{agree}$, their consensus result is returned. Otherwise, it proceeds to $k_{slow}$ (e.g., $\lceil k_{fast}/4 \rceil$) "slow thinking" verifications, and their consensus is returned.
    \item \textbf{\texttt{FormatInitialPrompt}}, \textbf{\texttt{FormatRetryPromptWithFeedback}}, \textbf{\texttt{ExtractFinalAnswer}}, \textbf{\texttt{MajorityVote}}: Standard utility functions for formatting prompts (see Figure~\ref{fig:proposer-prompt_appendix_v2} for initial prompt and Figure~\ref{fig:retry-prompt_appendix_v2} for retry prompt), extracting answers, and performing majority voting.
\end{itemize}

\paragraph{Solve}
Given a math problem \(x\), we employ DeepSeek‑R1‑14B as a step‑by‑step solution proposer (the LLM in \texttt{GenerateSolutionWithDetection}) using an initial prompt like the one in Figure \ref{fig:proposer-prompt_appendix_v2}. The prompt is sent in streaming chat‑completion mode, and tokens are appended sequentially to a buffer. If the initial solution attempt requires refinement based on verification feedback, a retry prompt like the one in Figure \ref{fig:retry-prompt_appendix_v2} is used.

\begin{figure*}[h]
\begin{minipage}{\textwidth}
\begin{tcolorbox}[
    colback=blue!5!white,
    colframe=blue!75!black,
    title=LLM Initial Solver Prompt,
    width=\textwidth, 
    enhanced,
    boxrule=1pt,
    arc=4mm,
    auto outer arc,
    fonttitle=\bfseries
]
\footnotesize
\begin{verbatim}
The following is a math problem:
[Math Problem]
{question}
Solve it step by step. For each step, you should use \n\n in the end.
Please put your final answer (i.e., the index) in \\boxed{{}}.
\end{verbatim}
\end{tcolorbox}
\end{minipage}
\caption{LLM Initial Solver Prompt (Appendix). This prompt structure is utilized by the \texttt{FormatInitialPrompt} helper function.} 
\label{fig:proposer-prompt_appendix_v2} 
\end{figure*}

\begin{figure*}[h]
\begin{minipage}{\textwidth}
\begin{tcolorbox}[
    colback=red!5!white, 
    colframe=red!75!black,
    title=LLM Retry Prompt with Feedback (Guided Solver),
    width=\textwidth, 
    enhanced,
    boxrule=1pt,
    arc=4mm,
    auto outer arc,
    fonttitle=\bfseries
]
\footnotesize
\begin{verbatim}
The following is a math problem:
[Math Problem]
{question}

You previously attempted to solve this problem, and your solution was:
[Previous Solution]
{previous_solution_S1}

That solution was reviewed, and the feedback is:
[Verification Feedback]
{verifier_feedback_F1}

Please carefully consider the feedback and correct your solution. \\
Provide a complete, new solution with clear reasoning steps. \\
Please put your final answer (i.e., the index) in \\boxed{{}}.
\end{verbatim}
\end{tcolorbox}
\end{minipage}
\caption{LLM Retry Prompt with Feedback (Appendix). This prompt structure is utilized by the \texttt{FormatRetryPromptWithFeedback} helper function. Placeholders such as \{question\}, \{previous\_solution\_S1\}, and \{verifier\_feedback\_F1\} are dynamically populated.} 
\label{fig:retry-prompt_appendix_v2} 
\end{figure*}

\paragraph{Detect}
The \texttt{GenerateSolutionWithDetection} function incorporates a mechanism to detect hesitation during reasoning. LLMs often employ hesitation words (e.g., "hmm", "let me verify") to self-verify. We observe that models may continue generating redundant checks even after reaching a solution. To decide when to truncate these overthinking situations and reduce redundant tokens, we use a streaming detection framework.

We first define the set of hesitation cues, shown in Figure \ref{fig:hesitation_keywords_appendix_v2}.

\begin{figure*}[h]
\begin{minipage}{\textwidth}
\begin{tcolorbox}[
    colback=blue!5!white,
    colframe=blue!75!black,
    title=LLM Detection Prompt,
    width=\textwidth, 
    enhanced,
    boxrule=1pt,
    arc=4mm,
    auto outer arc,
    fonttitle=\bfseries
]
\footnotesize
\begin{verbatim}
  Wait, double-check, Alternatively, Hmm, Let me check, 
  Alright, make sure, Another way, Let me verify, to confirm, 
  Looking back, But wait
\end{verbatim}
\end{tcolorbox}
\end{minipage}
\caption{A representative set of hesitation keywords monitored in the reasoning trace to detect potential solution completion (Appendix). (Requires \texttt{customtakeaway} environment definition.)}
\label{fig:hesitation_keywords_appendix_v2} 
\end{figure*}

As each token $t$ arrives during solution generation, if the end of the current reasoning sequence matches any hesitation keyword $k\in\mathcal{K}$ (where $\mathcal{K}$ is the set from Figure~\ref{fig:hesitation_keywords_appendix_v2}), we suspend the primary LLM proposer and trigger a detection process. A Detector LLM (which can be the same base model with a specific prompt) evaluates the current reasoning context using the prompt in Figure \ref{fig:detect-prompt_appendix_v2} to check whether a complete solution (including the final answer) has been reached. For efficiency, the Detector LLM is prompted to respond with only one token ("Yes" or "No") and minimal internal thought, for example:
\[
\langle\!\texttt{think}\rangle\;\texttt{Okay, I think I have finished thinking.}\;\langle\!/ \!\texttt{think}\rangle
\]

\begin{figure*}[h]
\begin{minipage}{\textwidth}
\begin{tcolorbox}[
    colback=blue!5!white,
    colframe=blue!75!black,
    title=LLM Detection Prompt,
    width=\textwidth, 
    enhanced,
    boxrule=1pt,
    arc=4mm,
    auto outer arc,
    fonttitle=\bfseries
]
\footnotesize
\begin{verbatim}
You are a solution completeness checker.
Given current solution to a math problem, determine if it is a complete
solution (i.e., contains a final answer).
Respond with exactly one word: `Yes` if complete, `No` otherwise.
\end{verbatim}
\end{tcolorbox}
\end{minipage}
\caption{LLM Detection Prompt (Appendix).} 
\label{fig:detect-prompt_appendix_v2} 
\end{figure*}

To improve decision robustness, we compare the log-probabilities of "Yes" and "No" from the Detector LLM's top token predictions. If \(\log p(\text{Yes})>\log p(\text{No})\), we conclude that the current reasoning contains a complete solution. If hesitation was detected and the solution deemed complete, the generation might be explicitly continued (as per the Python code's `continue-after-detected` logic) to capture any final utterances before concluding that segment of generation. The overall generation process then decides whether to terminate or proceed based on the pipeline's state.

\subsection{Scaling of \texttt{FlexiVe} Modes on ProcessBench}
\label{apdx:flexive_mode_scaling}

This section details the performance and token usage of \texttt{FlexiVe} when operating in its deliberative "With Thinking" (Think@k) mode versus its efficient "Without Thinking" (NoThinking@k) mode. The experiments were conducted on subsets of the ProcessBench benchmark (GSM8K, MATH, OlympiadBench, and OmniMATH) across various sampling budgets ($k$).

Table~\ref{tab:flexive_with_thinking_appendix_scaling} shows the F1 scores and total token consumption for \texttt{FlexiVe} in "With Thinking" mode. This mode generally achieves higher F1 scores, especially as the sampling budget $k$ increases, but at a significantly higher token cost.

\begin{table}[htbp]
\centering
\caption{Performance of \texttt{FlexiVe} "With Thinking" (Think@k) under different sampling budgets ($k$) on ProcessBench subsets. Tokens are total generated across the respective test set.}
\label{tab:flexive_with_thinking_appendix_scaling}
\resizebox{\textwidth}{!}{%
\begin{tabular}{@{}l cc cc cc cc@{}}
\toprule
& \multicolumn{2}{c}{GSM8K} & \multicolumn{2}{c}{MATH} & \multicolumn{2}{c}{OlympiadBench} & \multicolumn{2}{c}{OmniMATH} \\
\cmidrule(lr){2-3} \cmidrule(lr){4-5} \cmidrule(lr){6-7} \cmidrule(lr){8-9}
Voting Budget ($k$) & F1 (\%) & Tokens & F1 (\%) & Tokens & F1 (\%) & Tokens & F1 (\%) & Tokens \\
\midrule
2   & 82.3 & 2,412,972   & 81.9 & 5,209,255    & 78.0 & 8,428,333    & 71.3 & 7,055,913 \\
4   & 86.7 & 4,773,358   & 86.4 & 10,416,363   & 84.3 & 16,779,943   & 76.9 & 14,283,830 \\
8   & 86.4 & 9,534,029   & 88.9 & 20,913,932   & 85.4 & 33,417,171   & 78.9 & 28,633,370 \\
16  & 87.6 & 19,169,102  & 89.7 & 41,778,727   & 86.5 & 66,852,313   & 80.1 & 57,096,638 \\
32  & 87.7 & 38,055,768  & 89.7 & 83,807,676   & 86.7 & 133,587,678  & 80.6 & 114,215,045 \\
64  & 87.8 & 76,325,097  & 90.1 & 167,497,140  & 86.7 & 267,287,483  & 80.4 & 228,408,308 \\
128 & 88.1 & 152,675,054 & 90.0 & 335,401,726  & 86.7 & 534,138,821  & 80.5 & 456,401,199 \\
\bottomrule
\end{tabular}%
}
\end{table}

Conversely, Table~\ref{tab:flexive_without_thinking_appendix_scaling} presents the results for the "Without Thinking" mode. This mode is significantly more token-efficient, though it generally results in lower F1 scores compared to the "With Thinking" mode at equivalent sampling budgets. The trade-off between accuracy and computational cost is evident when comparing these two modes.

\begin{table}[htbp]
\centering
\caption{Performance of \texttt{FlexiVe} "Without Thinking" (NoThinking@k) under different sampling budgets ($k$) on ProcessBench subsets. Tokens are total generated across the respective test set.}
\label{tab:flexive_without_thinking_appendix_scaling}
\resizebox{\textwidth}{!}{%
\begin{tabular}{@{}l cc cc cc cc@{}}
\toprule
& \multicolumn{2}{c}{GSM8K} & \multicolumn{2}{c}{MATH} & \multicolumn{2}{c}{OlympiadBench} & \multicolumn{2}{c}{OmniMATH} \\
\cmidrule(lr){2-3} \cmidrule(lr){4-5} \cmidrule(lr){6-7} \cmidrule(lr){8-9}
Voting Budget ($k$) & F1 (\%) & Tokens & F1 (\%) & Tokens & F1 (\%) & Tokens & F1 (\%) & Tokens \\
\midrule
2   & 61.5 & 362,849     & 57.2 & 1,516,537    & 49.0 & 1,879,631    & 50.5 & 1,634,222 \\
4   & 66.8 & 737,332     & 61.3 & 3,040,918    & 53.8 & 3,725,119    & 52.5 & 3,317,988 \\
8   & 66.7 & 1,490,192   & 62.8 & 6,090,996    & 55.2 & 7,505,333    & 53.6 & 6,626,085 \\
16  & 66.8 & 2,973,364   & 64.3 & 12,107,246   & 55.9 & 15,025,214   & 54.2 & 13,258,722 \\
32  & 66.5 & 5,936,588   & 64.4 & 24,247,615   & 55.9 & 29,940,405   & 54.7 & 26,531,060 \\
64  & 66.8 & 11,833,305  & 64.2 & 48,501,840   & 56.1 & 59,802,922   & 54.0 & 52,945,921 \\
128 & 66.7 & 23,715,112  & 65.0 & 96,833,463   & 56.3 & 119,821,725  & 54.1 & 105,854,677 \\
\bottomrule
\end{tabular}%
}
\end{table}

\paragraph{Token Efficiency Summary}
The "Without Thinking" mode demonstrates substantial token savings compared to the "With Thinking" mode:
\begin{itemize}
    \item \textbf{GSM8K}: "Without Thinking" uses approximately 84.5\% fewer tokens.
    \item \textbf{MATH}: "Without Thinking" uses approximately 71.0\% fewer tokens.
    \item \textbf{OlympiadBench}: "Without Thinking" uses approximately 77.6\% fewer tokens.
    \item \textbf{OmniMATH}: "Without Thinking" uses approximately 76.8\% fewer tokens.
    \item \textbf{Average}: On average, the "Without Thinking" mode uses approximately 77.5\% fewer tokens than the "With Thinking" mode across these datasets.
\end{itemize}
This highlights the efficiency of the "NoThinking@k" approach for scenarios where computational budget is a primary constraint, while "Think@k" is preferable for achieving higher accuracy when more resources are available. The adaptive \texttt{FlexiVe} (Flex@k) mode, discussed in the main paper (Section~\ref{sec:ourmethod_performance}), aims to balance these two extremes.

\subsubsection{Supplementary Figures and Tables from Main Text Comments}

\begin{figure}[hbt!]
    \centering
    \includegraphics[width=0.75\columnwidth]{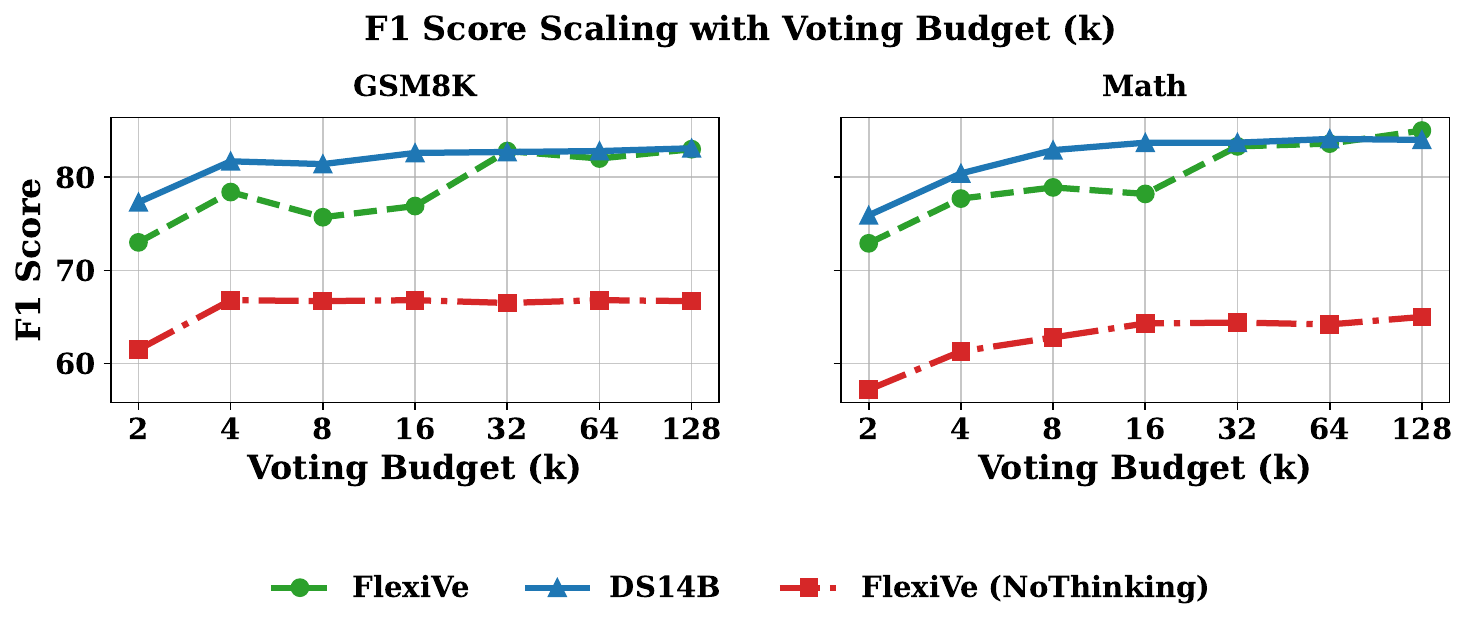} 
    \caption{F1 score scaling with voting budget $k$ on GSM8K (left) and MATH (right). \texttt{FlexiVe} (Flex@k, green circles) improves with larger $k$, performing comparably or better than DS14B (blue triangles, baseline verifier), while both surpass the \texttt{FlexiVe} (NoThinking variant, red squares). (Previously commented out from main text).}
    \label{fig:f1_scaling_appendix_v2}
\end{figure}

\begin{table*}[hbtp]
\centering
\caption{ProcessBench results reported with F1 scores. Results for \textbf{\ourmethod} are highlighted . \textbf{bold} indicates the best in the sub category. All \ourmethod~variants are trained on only 1526 samples.}
\resizebox{1.0\textwidth}{!}{
\begin{tabular}{lcccccc}
\toprule
\multirow{2}{*}{\textbf{Model}} & \multirow{2}{*}{\textbf{\# Samples}} & \multirow{2}{*}{\textbf{GSM8K}} & \multirow{2}{*}{\textbf{MATH}} & \multirow{2}{*}{\begin{tabular}[c]{@{}c@{}} \bf Olympiad \\ \bf Bench \end{tabular}} & \multirow{2}{*}{\begin{tabular}[c]{@{}c@{}} \bf Omni- \\ \bf MATH \end{tabular}} & \multirow{2}{*}{\textbf{Avg.}} \\
& \\
\midrule
\multicolumn{7}{c}{\textit{Proprietary Models}} \\
\midrule
GPT-4o-0806                      & unk          & 79.2          & 63.6          & 51.4          & 53.5          & 61.9 \\
o1-mini                           & unk          & 93.2          & 88.9          & 87.2          & 82.4          & 87.9 \\
\midrule
\multicolumn{7}{c}{\textit{Open Source Models (1.5B)}} \\
\midrule
Skywork-PRM-1.5B                  & unk          & 59.0          & 48.0          & 19.3          & 19.2          & 36.4 \\
\rowcolor{gray!10} \textcolor{gray}{GenPRM-1.5B (Pass@1) w/ Code Exec}      & \textcolor{gray}{23K}          & \textcolor{gray}{52.8}          & \textcolor{gray}{66.6}          & \textcolor{gray}{55.1}          & \textcolor{gray}{54.5}          & \textcolor{gray}{57.3} \\
\midrule
\multicolumn{7}{c}{\textit{Open Source Models (7-8B)}} \\
\midrule
Math-Shepherd-PRM-7B              & 445K         & 47.9          & 29.5          & 24.8          & 23.8          & 31.5 \\
RLHFlow-PRM-Mistral-8B            & 273K         & 50.4          & 33.4          & 13.8          & 15.8          & 28.4 \\
EurusPRM-Stage2                   & 30K          & 47.3          & 35.7          & 21.2          & 20.9          & 31.3 \\
Qwen2.5-Math-PRM-7B               & $\sim$344K   & 82.4          & 77.6          & 67.5          & 66.3          & 73.5 \\
RetrievalPRM-7B                   & 404K         & 74.6          & 71.1          & 60.2          & 57.3          & 65.8 \\
Universal-PRM-7B                  & unk          & 85.8          & 77.7          & 67.6          & 66.4          & 74.3 \\
Direct Generative PRM-7B          & 23K          & 63.9          & 65.8          & 54.5          & 55.9          & 60.0 \\
\rowcolor{gray!10} \textcolor{gray}{GenPRM-7B w/ Code Exec (Pass@1)}        & \textcolor{gray}{23K}          & \textcolor{gray}{78.7}          & \textcolor{gray}{80.3}          & \textcolor{gray}{72.2}          & \textcolor{gray}{69.8}          & \textcolor{gray}{75.2} \\
\rowcolor{gray!10} \textcolor{gray}{GenPRM-7B w/ Code Exec (Maj@8)}         & \textcolor{gray}{23K}          & \textcolor{gray}{81.0}          & \textcolor{gray}{85.7}          & \textcolor{gray}{78.4}          & \textcolor{gray}{76.8}          & \textcolor{gray}{80.5} \\
\midrule
\multicolumn{7}{c}{\textit{Open Source Models (14-32B) w/ \textbf{Moderate Compute}}} \\
\midrule
Dyve-14B                          & 117K         & 68.5          & 58.3          & 49.0          & 47.2          & 55.8 \\
GenPRM-32B w/o Code Exec (Maj@8)  & 23K          & 78.8          & \uline{85.1}    & 78.7          & \uline{74.9}    & 79.3 \\
\rowcolor{violet!10}\ourmethod (Flex@32)                     & \textbf{1526}         & \uline{82.8}    & 83.3          & \uline{79.2}    & 73.4          & \uline{79.7} \\
\rowcolor{violet!10}\ourmethod (Flex@128)                    & \textbf{1526}         & \textbf{83.0} & \textbf{85.0} & \textbf{80.0} & \textbf{75.2} & \textbf{80.8} \\
\midrule
\multicolumn{7}{c}{\textit{Open Source Models (14-32B) w/ \textbf{High Compute}}} \\
\midrule
\rowcolor{gray!10} GenPRM-32B (Pass@1) w/ Code Exec      & 23K          & 83.1          & 81.7          & 72.8          & 72.8          & 77.6 \\
\rowcolor{gray!10} GenPRM-32B (Maj@8) w/ Code Exec       & 23K          & 85.1          & 86.3          & 78.9          & 80.1          & 82.6 \\
\rowcolor{violet!10}\ourmethod (Think@64)                      & \textbf{1526}         & \textbf{88.1} & \textbf{90.1}  & \textbf{86.7}   & \textbf{80.4}             & \textbf{86.3}    \\
\bottomrule
\end{tabular}
}
\label{tab:processbench_results}%
\end{table*}%

\newpage

\end{document}